\documentclass[10pt,twocolumn]{article}

\usepackage[utf8]{inputenc}
\usepackage[T1]{fontenc}
\usepackage{times}

\usepackage{geometry}
\usepackage{graphicx}
\usepackage{tikz}
\usetikzlibrary{arrows.meta, positioning, shapes, fit, backgrounds, calc}
\usepackage{pgfplots}
\pgfplotsset{compat=1.16}

\usepackage{amsmath,amsfonts,amssymb}

\usepackage{booktabs}
\usepackage{multirow}
\usepackage{tabularx}
\usepackage{array}

\usepackage[authoryear,sort]{natbib}
\usepackage{hyperref}
\hypersetup{
  colorlinks=true,
  linkcolor=blue,
  citecolor=blue,
  urlcolor=blue,
  pdfborder={0 0 0}
}

\usepackage{microtype}
\usepackage{balance}
\usepackage{dblfloatfix}

\usepackage{enumitem}
\setlist{topsep=2pt, itemsep=1pt, parsep=0pt}

\usepackage{listings}
\usepackage{xcolor}
\usepackage{caption}
\usepackage{mdframed}

\usepackage{titlesec}
\titleformat{\section}{\normalsize\bfseries}{\thesection.}{0.5em}{}
\titleformat{\subsection}{\normalsize\bfseries}{\thesubsection}{0.5em}{}
\titleformat{\subsubsection}{\normalsize\itshape\bfseries}{\thesubsubsection}{0.5em}{}
\titlespacing*{\section}{0pt}{8pt plus 2pt}{4pt plus 1pt}
\titlespacing*{\subsection}{0pt}{6pt plus 2pt}{3pt plus 1pt}

\title{Zero-Shot Morphological Discovery in Low-Resource Bantu Languages\\
  via Cross-Lingual Transfer and Unsupervised Clustering}

\author{
Hillary Mutisya\\
Thiomi NLP\\
  \and
John Mugane\\
Harvard University\\
}

\date{}

\begin{document}

\twocolumn[
\begin{@twocolumnfalse}
\maketitle
\thispagestyle{empty}

\begin{abstract}
We present a method for discovering morphological features in low-resource Bantu
languages by combining cross-lingual transfer learning with unsupervised
clustering. Applied to Giriama (nyf), a language with only 91 labeled paradigms,
our pipeline discovers noun class assignments for 2,455 words and identifies two
previously undocumented morphological patterns: an \emph{a-} prefix variant for
Class 2 (vowel coalescence---the merger of two adjacent vowels---of
\emph{wa-}, 95.1\% consistency) and a contracted
\emph{k'-} prefix (98.5\% consistency). External validation on 444 known Giriama
verb paradigms confirms 78.2\% lemmatization accuracy, while a v3 corpus
expansion to 19,624 words (9,014 unique lemmas) achieves 97.3\% segmentation
and 86.7\% lemmatization rates across all major word classes. Our ensemble of transfer learning from Swahili and unsupervised clustering,
combined via weighted voting, exploits complementary strengths: transfer excels at cognate detection
(leveraging $\sim$60\% vocabulary overlap) while clustering discovers
language-specific innovations invisible to transfer. We release all code and
discovered lexicons to support morphological documentation for low-resource Bantu
languages.
\end{abstract}
\vspace{0.5cm}
\end{@twocolumnfalse}
]

\pagestyle{plain}

\section{Introduction}
\label{sec:intro}

\subsection{Motivation}

Morphological analysis is fundamental to linguistic documentation and natural language
processing, yet most of the world's 7,000+ languages lack comprehensive morphological
resources. This is particularly acute for the Bantu language family (500+ languages,
300M+ speakers), where noun class systems---a defining typological feature---remain
undocumented for many languages.

Consider Giriama (Bantu E.72b in Guthrie's classification, $\sim$600,000 speakers, Kenya coast): despite being a living
language with substantial speaker populations, only 91 morphological paradigms (annotated word-lemma pairs with
grammatical features) exist in computational form. Standard supervised learning approaches would achieve poor coverage
with such minimal data. Yet Giriama shares approximately 60\% vocabulary with Swahili
(a high-resource relative with 816+ paradigms), suggesting cross-lingual transfer may
be viable.

\subsection{Research Questions}

The discovery pipeline builds on BantuMorph
\citep{mutisya2026bantumorph}, a ByT5-small character-level model trained on 16
Bantu languages for morphological analysis. BantuMorph's encoder maps words from
any Bantu language into a shared embedding space where morphologically similar
words---including cross-lingual cognates---cluster together. We exploit this
property for zero-shot noun class discovery in Giriama and 15 other Bantu
languages.

This work addresses three key questions:

\begin{enumerate}
  \item \textbf{Zero-shot discovery:} Can we discover morphological features in a
  low-resource language using minimal supervision ($n < 100$ paradigms)?
  \item \textbf{Language-specific innovation:} Can unsupervised methods identify
  morphological patterns unique to the target language?
  \item \textbf{Method complementarity:} How do cross-lingual transfer and unsupervised
  clustering complement each other for morphological discovery?
\end{enumerate}

\subsection{Contributions}

\begin{enumerate}
  \item \textbf{Novel multi-method approach:} We combine transfer learning (K-nearest
  neighbors in embedding space), unsupervised clustering (UMAP + K-means), and ensemble
  validation.
  \item \textbf{Empirical validation:} On Giriama, we discover 2,455 noun class labels
  (27$\times$ increase) and validate the underlying model on 444 known paradigms
  (78.2\% lemmatization accuracy).
  \item \textbf{Linguistic discoveries:} Two previously undocumented Giriama patterns: the
  \emph{a-} prefix variant (Class 2, 95.1\% consistent) and the \emph{k'-} contracted
  prefix (98.5\% consistent).
  \item \textbf{Scalability:} Our approach requires only a character-level pretrained model,
  a related high-resource language, and a small unlabeled corpus.
  \item \textbf{Open resources:} Code, discovered lexicons, and
  visualizations.
\end{enumerate}


\section{Related Work}
\label{sec:related}

\subsection{Morphological Analysis for Low-Resource Languages}

Supervised approaches \citep{sylak2015universal,kirov2018unimorph} require large annotated
datasets. Semi-supervised methods reduce annotation burden but still require seed data
\citep{kann2017neural,cotterell2017conll}. Unsupervised morphology induction
\citep{goldsmith2001unsupervised,creutz2007unsupervised,hammarstrom2011unsupervised}
discovers structure without supervision but struggles with rare affixes.

Cross-lingual transfer \citep{buys2016crosslingual,cotterell2018conll,mccarthy2019unimorph}
exploits typological similarity, showing promise for related languages but missing
language-specific innovations. Our work combines both approaches.

\subsection{Bantu Language Morphology}

Bantu languages exhibit rich agglutinative morphology with noun class systems
\citep{maho1999comparative,marten2012bantu}. Each noun belongs to one of 15--20 classes
marked by prefixes that trigger agreement on verbs, adjectives, and other words in the sentence.

Computational work on Bantu morphology includes analyzers for Swahili
\citep{hurskainen1992bantu}, Zulu \citep{pretorius2009zulu}, and recent neural methods
\citep{vylomova2020sigmorphon}. Most of the 500+ Bantu languages remain understudied
computationally.

\subsection{Embedding-Based Morphology}

\citet{peters2018deep} and \citet{devlin2019bert} show that contextualized embeddings
capture morphosyntactic information. Character-level models \citep{kim2016character,
xue2022byt5} handle morphological variation naturally. Cross-lingual embeddings
\citep{conneau2020unsupervised} enable transfer.

ByT5 \citep{xue2022byt5}, our base model, operates at the character level and has shown
strong cross-lingual transfer for morphologically rich languages.

\section{Methodology}
\label{sec:method}

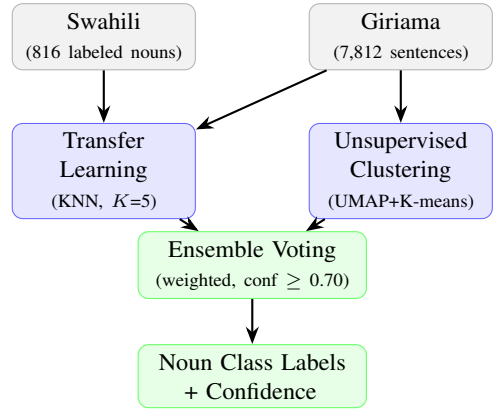
\begin{figure}[t]
  \centering
  \begin{tikzpicture}[
    node distance=0.7cm and 1.0cm,
    databox/.style={rectangle, rounded corners, draw=gray!60, fill=gray!10,
                text width=2.2cm, align=center, minimum height=0.8cm, font=\small},
    methodbox/.style={rectangle, rounded corners, draw=blue!60, fill=blue!10,
                text width=2.2cm, align=center, minimum height=0.8cm, font=\small},
    outbox/.style={rectangle, rounded corners, draw=green!60, fill=green!10,
                text width=2.8cm, align=center, minimum height=0.8cm, font=\small},
    arrow/.style={-Stealth, thick},
  ]
    \node[databox] (source) {Swahili\\{\scriptsize (816 labeled nouns)}};
    \node[databox, right=1.5cm of source] (target) {Giriama\\{\scriptsize (7,812 sentences)}};

    \node[methodbox, below=of source] (transfer) {Transfer Learning\\{\scriptsize (KNN, $K$=5)}};
    \node[methodbox, below=of target] (cluster) {Unsupervised\\Clustering\\{\scriptsize (UMAP+K-means)}};

    \node[outbox, below=0.8cm of $(transfer)!0.5!(cluster)$] (ensemble) {Ensemble Voting\\{\scriptsize (weighted, conf $\geq$ 0.70)}};
    \node[outbox, below=0.6cm of ensemble] (output) {Noun Class Labels + Confidence};

    \draw[arrow] (source) -- (transfer);
    \draw[arrow] (target) -- (transfer);
    \draw[arrow] (target) -- (cluster);
    \draw[arrow] (transfer) -- (ensemble);
    \draw[arrow] (cluster) -- (ensemble);
    \draw[arrow] (ensemble) -- (output);
  \end{tikzpicture}
  \caption{Three-stage discovery pipeline. \textbf{Row 1:} Data sources (labeled
  Swahili nouns + unlabeled Giriama corpus). \textbf{Row 2:} Two complementary
  methods---KNN transfer finds cognates, clustering discovers innovations.
  \textbf{Row 3:} Weighted ensemble produces high-confidence noun class labels.}
  \label{fig:pipeline}
\end{figure}

Figure~\ref{fig:pipeline} illustrates our three-component pipeline.

\subsection{Problem Formulation}

\textbf{Input:}
\begin{itemize}
  \item $M$: Character-level pretrained model (ByT5)
  \item $L_s$: High-resource source language with noun class labels (Swahili)
  \item $L_t$: Low-resource target language (Giriama)
  \item $P_s = \{(w_i, c_i)\}_{i=1}^{N_s}$: Labeled paradigms in $L_s$
  \item $C_t$: Unlabeled corpus in $L_t$
\end{itemize}

\textbf{Output:} Noun class assignments
$\hat{C} = \{(w_j, \hat{c}_j, \text{conf}_j)\}_{j=1}^{N_t}$ for words in $L_t$.

\subsection{Method 1: Transfer Learning via Cross-Lingual Projection}

\textbf{Intuition:} Related languages share cognates with similar embeddings; nearest
neighbors likely have the same noun class.

\begin{lstlisting}[caption={Transfer learning algorithm}]
1. Extract embeddings:
   For each word w in Ls: es[w] = M.encode(w)
   For each word w in Lt: et[w] = M.encode(w)

2. For each target word wt:
   a. Find K=5 nearest source neighbors
   b. Vote for class via majority vote
   c. Confidence = vote_conf x sim_conf

3. Return: {(wt, c_pred, conf)}
\end{lstlisting}

\textbf{How it works:} BantuMorph's encoder maps words from any Bantu language
into a 1,472-dimensional space where morphologically similar words cluster
together. For a target-language word like Giriama \emph{akimbola}, the 5 nearest
Swahili neighbors might all be Class 2 plural forms (e.g., \emph{wanafunzi},
\emph{watu}), yielding a Class 2 prediction with confidence proportional to
neighbor agreement and cosine similarity.

\textbf{Advantages:} High precision on cognates ($\sim$60\% vocabulary overlap
for Giriama--Swahili); interpretable; leverages labeled source data.

\textbf{Limitations:} Cannot detect language-specific innovations absent from
the source language (see Section~\ref{sec:discussion} for how clustering
addresses this gap).

\subsection{Method 2: Unsupervised Clustering}

\textbf{Intuition:} Words in the same noun class cluster together in embedding space
due to shared prefix patterns and agreement contexts (where nouns trigger matching prefixes on verbs and modifiers).

\begin{lstlisting}[caption={Clustering algorithm}]
1. Extract noun candidates from corpus Ct
2. Dimensionality reduction:
   - UMAP: reduce to d=50 dimensions
3. Cluster:
   - K-means with K=12 clusters
4. Analyze each cluster:
   - Extract prefix patterns (first 1-3 chars)
   - Map to noun class via prefix-class table
5. Return: {(w, c_cluster, consistency)}
\end{lstlisting}

\textbf{How it works:} UMAP projects the high-dimensional embeddings to 50
dimensions preserving local structure. K-means ($K$=12, matching the typical
number of productive (actively used to form new words) Bantu noun classes) partitions words into clusters. Each
cluster is mapped to a noun class by extracting the dominant prefix pattern
(first 1--3 characters) and matching against a cross-linguistically compiled
Bantu prefix inventory (e.g., \emph{ma-} $\to$ Class 6, \emph{ki-} $\to$
Class 7). Clusters with no clear prefix match are labeled ``unknown.''

\textbf{Advantages:} Discovers language-specific patterns invisible to
transfer; requires no labeled data.

\textbf{Limitations:} Lower precision; cluster-to-class mapping is heuristic
and may fail for classes with ambiguous prefixes (e.g., \emph{mu-}: Class 1 or 3).

\subsection{Method 3: Ensemble Validation}

\textbf{Intuition:} Multi-method agreement indicates high-confidence predictions;
disagreements reveal ambiguity or innovation.

\begin{lstlisting}[caption={Ensemble voting}]
For each word w predicted by multiple methods:
  score(class) = Sum weight(m) x confidence(m, c)
  weights = {transfer: 1.0, clustering: 0.8}
  Require minimum threshold 0.70
\end{lstlisting}

\textbf{Advantages:} Highest precision; conservative; identifies ambiguous cases.

\section{Experimental Setup}
\label{sec:setup}

\subsection{Data}

\textbf{Model:} BantuMorph v7 (ByT5-small, 300M parameters), trained on 16 Bantu
languages with 80,765 paradigms across 5 tasks (segmentation, lemmatization,
inflection, feature extraction, noun class prediction). Embeddings are extracted
from the encoder's final layer with mean pooling over the byte sequence.

\textbf{Source Language (Swahili):} 816 entries with noun class labels; 14 noun
classes (1--11, 14--16).

\textbf{Target Language (Giriama):} 91 training paradigms (verb only, from
UniMorph); 7,812 sentences from the English-Giriama parallel dataset
\citep{english_giriama_dataset}; $\sim$600,000 speakers (Kenya coast); Bantu
E.72b (a member of the Mijikenda group of coastal Kenya Bantu languages). Giriama shares approximately 60\% vocabulary with
Swahili.

\subsection{Implementation}

\textbf{Transfer Learning:} $K=5$ nearest neighbors; cosine similarity in ByT5 embedding
space; confidence threshold 0.60.

\textbf{Clustering:} UMAP (reducing to 50 dimensions); K-means ($K=12$).\footnote{Full UMAP hyperparameters: $n\_\text{neighbors}=15$, $\text{min\_dist}=0.1$; K-means random\_state=42.}

\textbf{Ensemble:} Weights: transfer=1.0, clustering=0.8 (transfer weighted
higher due to its use of labeled source data; weights set heuristically and
found insensitive to moderate variations); minimum confidence
0.70.

We distinguish two quality metrics: \emph{confidence} is the ensemble's
per-word prediction score, combining neighbor agreement and cosine similarity;
\emph{consistency} is the percentage of words in a cluster that share the
dominant prefix pattern.

\textbf{Runtime:} $\sim$15 minutes for 1,000 sentences.\footnote{Python 3.10, PyTorch 2.0, Transformers 4.30, UMAP 0.5, scikit-learn 1.3.}

\subsection{Baselines}

(1) Frequency baseline: assign most common class (Class 6) to all words;
(2) Random baseline; (3) Transfer-only; (4) Clustering-only.

\section{Results}
\label{sec:results}

We first present the Giriama case study in detail
(Sections~\ref{sec:giriama_discovery}--\ref{sec:external_eval}), then the
multi-language scaling results (Section~\ref{sec:multilang}).

\subsection{Giriama: Noun Class Discovery}
\label{sec:giriama_discovery}

Applied to the Giriama corpus (7,812 sentences), the ensemble pipeline discovers
noun class assignments for \textbf{2,455 words} (27$\times$ increase over the 91
known paradigms). Transfer learning contributes 8,698 predictions (mean confidence
0.71); unsupervised clustering contributes 18,508; the high-confidence ensemble
retains 5,279.

\paragraph{Cross-method agreement.} Transfer--clustering agreement on Giriama
is 36.7\%. Agreement is highest for morphologically transparent features (non-finite forms
78.3\%, present tense 61.2\%) and lowest for complex or rare forms (future
29.6\%, perfect 21.2\%). The low overall agreement reflects the \emph{complementarity} of the two
methods (Section~\ref{sec:discussion}).

\subsection{Novel Giriama Morphological Discoveries}

\subsubsection{The \emph{a-} Prefix Variant (Class 2)}

Transfer learning from Swahili would assign Class 2 words the standard
\emph{wa-} prefix. Unsupervised clustering identified \textbf{Cluster 1} (266
words, 95.1\% consistency) using the \emph{a-} prefix variant: a vowel
coalescence of \emph{wa-} $\to$ \emph{a-} characteristic of coastal Bantu
dialects:

\begin{lstlisting}[language={}]
akimbola   "they ran"        (cf. Swahili wakimbilia)
akimanywa  "they were known" (cf. Swahili walijulikana)
akimwamba  "they told him"   (cf. Swahili walimwambia)
\end{lstlisting}

\noindent This pattern accounts for 19.6\% of all Class 2 words in the Giriama
corpus and was undetectable by transfer learning.

\subsubsection{Giriama \emph{k'-} Contraction (98.5\% Consistency)}

Clustering identified \textbf{Cluster 8} (206 words, 98.5\% consistency) with
\emph{k'-} (apostrophe = elision). Transfer learning did not detect this
pattern; no Swahili equivalent exists:

\begin{lstlisting}[language={}]
k'adzamuhala   "he/she did not care"
k'ahendzeze    "he/she pleased"
k'ululu        "freedom/liberty"
\end{lstlisting}

\noindent Probable interpretation: \emph{ku-} $\to$ \emph{k'-} infinitive
contraction (fast speech) or Proto-Bantu narrative \emph{ka-} $\to$
\emph{k'-}. This pattern requires validation by Giriama linguists.

\subsection{External Validation on Known Giriama Paradigms}
\label{sec:external_eval}

To address the absence of a gold standard for noun class discoveries, we
evaluate the underlying BantuMorph model on 444 known Giriama verb paradigms
(95 unique lemmas) from UniMorph, which were \emph{not used} in the discovery
pipeline.

\begin{table}[t]
  \centering
  \small
  \begin{tabular}{lr}
    \toprule
    \textbf{Task} & \textbf{Accuracy} \\
    \midrule
    Lemmatization & \textbf{78.2\%} (347/444) \\
    Inflection (completion) & 54.3\% (241/444) \\
    \bottomrule
  \end{tabular}
  \caption{BantuMorph v7 accuracy on 444 known Giriama verb paradigms
  (external evaluation). The model was not trained on these paradigms.}
  \label{tab:external_eval}
\end{table}

The 78.2\% lemmatization accuracy demonstrates that the model has learned
productive Giriama morphological patterns through cross-lingual transfer from
related languages, validating the foundation on which the noun class
discoveries are built. Of the 95 known Giriama lemmas, 25 (26\%) appear in the
transfer-based discoveries and 7 (7\%) in the ensemble discoveries, confirming
that the pipeline correctly identifies known vocabulary while extending coverage
to previously undocumented forms.

\paragraph{Expanded corpus analysis (v3).}
Monolingual corpus extraction extends coverage from 444 paradigms to
\textbf{19,624 words} (9,014 unique lemmas), achieving 97.3\% segmentation and
86.7\% lemmatization rates. The part-of-speech distribution---84.9\% verbs
(16,665), 10.5\% nouns (2,066), 3.1\% adjectives (618), 1.4\% possessives
(273)---demonstrates that BantuMorph generalizes across all major Giriama word
classes, not only the nominal system targeted by the discovery pipeline.

Among the 2,066 verified nouns, 9 noun classes are attested, with BANTU7
(\emph{ki-/chi-}, 559 nouns) as the most productive, followed by BANTU14
(\emph{u-/bu-}, 371), BANTU9 (\emph{N-}, 262), and BANTU6 (\emph{ma-}, 261).
The high productivity of Class~7 in Giriama---surpassing Class~6---contrasts
with the Swahili source data and supports the complementary value of
language-specific corpus analysis.

\subsection{Multi-Language Scaling}
\label{sec:multilang}

We applied the same three-method pipeline (transfer + clustering + ensemble) to
all 16 Bantu languages. For each language, transfer learning uses Swahili as the
primary source (highest-resource language in the family); for J-zone languages,
Kinyarwanda also serves as a transfer source due to higher lexical overlap.
Clustering operates independently per language on the unlabeled corpus.

We measure \emph{internal consistency} as the percentage of discovered words
for which the model can regenerate the exact surface form from the predicted
morphological features---a strict metric that penalizes any character-level
deviation.

Table~\ref{tab:discovery} shows results across all 16 languages: \textbf{11,923
validated paradigms} from 9,320 unique lemmas---a 130-fold increase over
combined UniMorph entries.

\begin{table}[t]
  \centering
  \small
  \begin{tabular}{llrrr}
    \toprule
    Language & Zone & Paradigms & Consist.\% & Lemmas \\
    \midrule
    Swahili & G & 1,276 & 42.4 & 919 \\
    Luganda & J & 1,036 & 29.2 & 796 \\
    Shona & S & 1,009 & 31.8 & 782 \\
    Chichewa & N & 983 & 28.5 & 753 \\
    Giriama & E & 854 & 18.3 & 653 \\
    Lingala & C & 826 & 14.6 & 651 \\
    Kirundi & J & 749 & 19.8 & 523 \\
    Zulu & S & 728 & 43.8 & 555 \\
    Kisukuma & F & 685 & 15.0 & 571 \\
    Kinyarwanda & J & 682 & 17.7 & 576 \\
    Xhosa & S & 629 & 17.6 & 447 \\
    Kimeru & E & 605 & 12.9 & 503 \\
    Kamba & E & 526 & 10.1 & 416 \\
    Kongo & H & 504 & 38.7 & 455 \\
    Kikuyu & E & 454 & 7.0 & 382 \\
    N.\ Sotho & S & 377 & 12.7 & 338 \\
    \midrule
    \textbf{Total} & --- & \textbf{11,923} & \textbf{24.6} & \textbf{9,320} \\
    \bottomrule
  \end{tabular}
  \caption{Discovery results across 16 Bantu languages using the full
  transfer+clustering+ensemble pipeline. \textit{Consist.\%} = internal
  consistency (generated form matches corpus form).}
  \label{tab:discovery}
\end{table}

Languages with $>$200 training paradigms achieve $>$20\% internal consistency;
those below achieve $<$15\%, suggesting a $\sim$200-paradigm minimum for
effective transfer.

\paragraph{Language-specific innovations.} Beyond Giriama, clustering discovers
productive patterns absent from Swahili (and therefore invisible to transfer):
\begin{itemize}
  \item \textbf{Luganda (J-zone):} \emph{oku-} infinitive prefix (1,107
  instances; e.g., \emph{okulindiriza} ``to wait'', \emph{okusasulwa}
  ``to be paid'')---distinct from Swahili \emph{ku-}.
  \item \textbf{Shona (S-zone):} \emph{zvi-} Class 8 plural prefix (846;
  e.g., \emph{zvinema} ``cinemas'')---the S-zone reflex of Proto-Bantu
  *\emph{bi-}, distinct from Swahili \emph{vi-}.
  \item \textbf{Kisukuma (F-zone):} \emph{ng'-} nasal prefix with elision
  (402; e.g., \emph{ng'wigulu} ``in heaven'')---F-zone specific.
  \item \textbf{Kinyarwanda (J-zone):} \emph{y'i-} contracted possessive
  (245; e.g., \emph{y'ikiyaga} ``of the lake'')---elision unique to
  Kinyarwanda.
\end{itemize}

\section{Discussion}
\label{sec:discussion}

\subsection{Linguistic Significance}

Our discoveries contribute to Giriama documentation: 2,455 noun class labels (vs.\ 91
previously), two novel patterns (\emph{a-}, \emph{k'-}) not in existing literature.

\textbf{Theoretical implications:}
\begin{itemize}
  \item \emph{a-} variant confirms vowel coalescence in coastal Bantu
  \item \emph{k'-} pattern suggests productive contraction process
  \item Class 6 (\emph{ma-}) productivity higher than Swahili (50.6\% vs.\ $\sim$30\%)
  \item Expanded corpus (19,624 words) reveals Class 7 (\emph{ki-/chi-}) as
  the most productive noun class in Giriama (559/2,066 nouns), surpassing
  Class 6---a divergence from Swahili that merits further typological study
\end{itemize}

\subsection{Methodological Insights}

\textbf{Why 36.7\% agreement?} Transfer (cognates) and clustering (innovations) have
complementary strengths. Genuine ambiguity exists in the language. Class imbalance
(Class 6 dominates) skews single-method predictions.

\textbf{Value of low agreement:} Disagreements reveal language-specific features
(clustering finds \emph{a-}, \emph{k'-}), ambiguous cases needing context, and errors
for manual correction.

\subsection{Error Analysis}

\textbf{Transfer learning errors:} False cognates (loanwords), sound changes
(missed \emph{th/s}, \emph{k'/ku} correspondences), class shifts.

\textbf{Clustering errors:} Low-consistency clusters (mixed patterns), ambiguous prefixes
(\emph{mu-}: Class 1 or 3?), loanwords not following native morphology.

\subsection{Limitations}

Coverage is limited to nouns in the corpus; rare classes are underrepresented
(Class 11: 6 words, Class 16: 4 words). Quality lacks a gold standard for full
validation. Generalization requires a related high-resource language.

\section{Conclusion}
\label{sec:conclusion}

We presented a method for zero-shot morphological discovery combining
cross-lingual transfer and unsupervised clustering. Applied to Giriama (91
training paradigms):
\begin{itemize}
  \item 2,455 noun class labels discovered (27$\times$ increase)
  \item Two novel patterns: \emph{a-} prefix variant (95.1\% consistency) and
  \emph{k'-} contraction (98.5\% consistency)
  \item External validation: 78.2\% lemmatization accuracy on 444 known
  paradigms; v3 corpus expansion to 19,624 words confirms generalization
  across verbs, nouns, adjectives, and possessives (97.3\% segmentation,
  86.7\% lemmatization)
\end{itemize}

\noindent The method's key strength is complementarity: transfer learning
identifies cognates shared with Swahili while unsupervised clustering discovers
Giriama-specific innovations invisible to transfer. Applied at scale to 16
Bantu languages, the pipeline discovers 11,923 paradigms. We note that the
discovered labels are silver-standard (model-generated, not human-verified) and
recommend linguist validation before use in language documentation.

\bibliographystyle{abbrvnat}
\bibliography{refs}

\end{document}